\documentclass[letterpaper]{article} 
\usepackage{aaai2026}  
\usepackage{times}  
\usepackage{helvet}  
\usepackage{courier}  
\usepackage[hyphens]{url}  
\usepackage{graphicx} 

\usepackage{natbib}  
\usepackage{caption} 
\usepackage{algorithm}
\usepackage{algorithmic}
\usepackage{amssymb}
\usepackage{amsmath}
\usepackage{multirow}
\usepackage{newfloat}
\usepackage{listings}
\DeclareCaptionStyle{ruled}{labelfont=normalfont,labelsep=colon,strut=off} 
\floatstyle{ruled}
\newfloat{listing}{tb}{lst}{}
\floatname{listing}{Listing}
\title{Spatiotemporal-Untrammelled Mixture of Experts for Multi-Person Motion Prediction}
\author {
    Zheng Yin\textsuperscript{\rm 1}\equalcontrib,
    Chengjian Li\textsuperscript{\rm 1}\equalcontrib,
    Xiangbo Shu\textsuperscript{\rm 1}\thanks{Corresponding author.}, 
    Meiqi Cao\textsuperscript{\rm 1}, 
    Rui Yan\textsuperscript{\rm 1}, 
    Jinhui Tang\textsuperscript{\rm 2}
}
\affiliations {
    \textsuperscript{\rm 1}Nanjing University of Science and Technology\\
    \textsuperscript{\rm 2}Nanjing Forestry University\\
   alanyz@njust.edu.cn, lichengjian@njust.edu.cn, shuxb@njust.edu.cn
}

\usepackage{bibentry}

\begin{document}

\maketitle

\begin{abstract}
Comprehensively and flexibly capturing the complex spatio-temporal dependencies of human motion is critical for multi-person motion prediction. 
Existing methods grapple with two primary limitations: i) Inflexible spatiotemporal representation due to reliance on positional encodings for capturing spatiotemporal information. ii) High computational costs stemming from the quadratic time complexity of conventional attention mechanisms. To overcome these limitations, we propose the Spatiotemporal-Untrammelled Mixture of Experts (ST-MoE), which flexibly explores complex spatio-temporal dependencies in human motion and significantly reduces computational cost. To adaptively mine complex spatio-temporal patterns from human motion, our model incorporates four distinct types of spatiotemporal experts, each specializing in capturing different spatial or temporal dependencies.
To reduce the potential computational overhead while integrating multiple experts, we introduce bidirectional spatiotemporal Mamba as experts, each sharing bidirectional temporal and spatial Mamba in distinct combinations to achieve model efficiency and parameter economy. Extensive experiments on four multi-person benchmark datasets demonstrate that our approach not only outperforms state-of-art in accuracy but also reduces model parameter by 41.38\% and achieves a 3.6× speedup in training. The code is
available at https://github.com/alanyz106/ST-MoE.
\end{abstract}


\section{Introduction}

Human motion prediction aims to forecast future human movements from observed motion sequences. The field carries substantial importance for applications including human-robot interaction \cite{gui2018adversarial,zhuo2019unsupervised,jiang2024delving}, autonomous driving \cite{tang2023collaborative,fang2023tbp,cao2025exploiting,jiang2025fine}, and surveillance systems \cite{vu2020anomaly,qu2025learning,qu2025mvp,qu2025omnigaze,xing2025vision}. By analyzing historical human motion, robotic systems can infer human intentions, enabling more effective collaboration. Although traditional approaches focus on single individuals \cite{butepage2017deep,mao2019learning,cui2020learning,shu2021spatiotemporal}, multi-person motion prediction (MPMP) holds greater practical relevance, as real-world scenarios typically involve multiple individuals. Recent multi-person motion prediction methods leverage Transformer to learn spatial relationships between joints via self-attention \cite{xu2023joint}, neglecting the importance of capturing spatiotemporal dependencies in multi-person motion.

\begin{figure*}[t]
\centering
\includegraphics[width=0.985\textwidth]{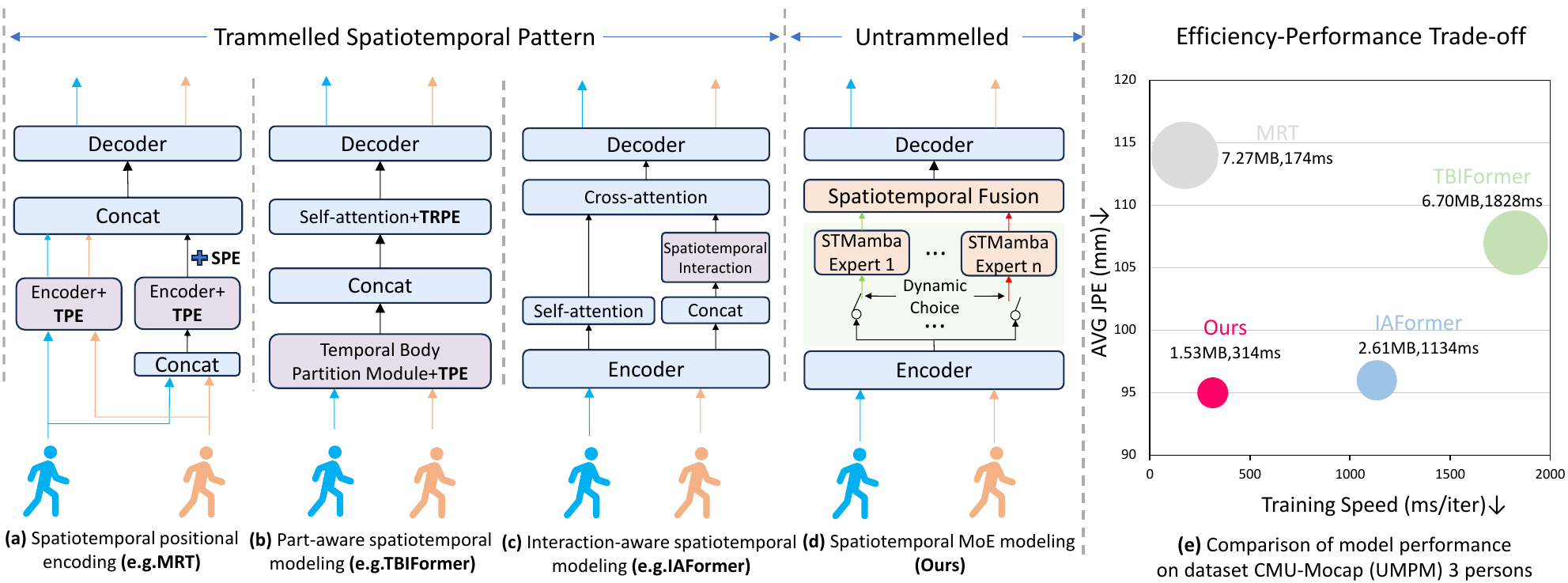} 
\caption{Insight of our work and performance of models on dataset CMU-Mocap(UMPM). (a)-(c): Limitations of prior methods using attention mechanism/spatio-temporal positional encodings, which solely capture trammelled spatiotemporal patterns. \textbf{TPE}/\textbf{SPE} denote temporal/spatial positional encoding. \textbf{TRPE} denotes trajectory-aware relative position encoding. (d): Our MoE-based framework with dynamic expert selection for adaptive spatiotemporal modeling. (e): Our model achieves an optimal efficiency-performance trade-off. For fair comparison, the batch size of all models was set to 96.}
\label{fig1}
\end{figure*}

To this end, MRT \cite{guo2022multi} employs temporal positional encoding to capture changes in human posture over time, and uses spatial positional encoding in the global encoder to enhance spatial correlations (Fig. \ref{fig1}(a)). However, this spatiotemporal position encoding employs a fixed pattern and lacks flexibility, leading to constrained prediction performance \cite{chu2021conditional}. TBIFormer \cite{peng2023trajectory} partitions human body parts and incorporates trajectory-aware relative position encoding, thereby enhancing the model's spatial perception (Fig. \ref{fig1}(b)). However, due to the quadratic computational complexity of the attention mechanism, the body part concatenation operation increases sequence length, significantly elevating computational costs. On the other hand, IAFormer \cite{xiao2024multi} utilizes self-attention mechanisms to explore spatiotemporal features within interactive information, significantly improving performance (Fig. \ref{fig1}(c)). Yet like TBIFormer, this method remains constrained by the efficiency bottleneck.

Despite the success of Transformer-based methods, these approaches continue to exhibit two key limitations: (a) Constrained model architectures fail to flexibly and adequately explore the complex spatio-temporal dependencies inherent in human motion. (b) Excessive dependence on computationally intensive attention mechanisms results in suboptimal efficiency. One question is naturally motivated to ask: \textit{Can we devise a new efficient paradigm for multi-person motion that flexibly and comprehensively captures spatio-temporal dependencies in human movement?}

To answer this question, we propose a lightweight framework \textbf{ST-MoE} for efficient and flexible modeling of complex spatiotemporal dependencies in multi-person motion prediction (Fig. \ref{fig1}(d)). Specifically, inspired by the dynamic activation mechanism of subnetworks in Mixture-of-Experts (MoE), we design four heterogeneous experts that respectively model distinct spatiotemporal patterns. Unlike conventional approaches relying on attention mechanisms or explicit positional encodings, ST-MoE adaptively activates optimal experts based on varied spatiotemporal features, thereby enhancing flexible modeling for diverse motion.
To improve modeling efficiency, we replace the traditional high-cost spatiotemporal attention mechanism with the Mamba, which has linear complexity. Specifically, ST-MoE flexibly configures combinations of bidirectional spatial and temporal Mamba for each expert to capture diverse spatiotemporal interaction patterns, and facilitate cross-dimensional feature fusion and parameter compression by sharing spatiotemporal Mamba.
Under the synergy of MoE's dynamic routing mechanism and Mamba's efficient modeling, ST-MoE achieves a significant reduction in computational cost while maintaining prediction accuracy.
Results on CMU-Mocap (UMPM) \cite{van2011umpm,CMUGraphicsLabMocap2003} confirm our approach's superiority over the SOTA IAFormer \cite{xiao2024multi} in prediction accuracy, while simultaneously accelerating training by 3.6× and reducing model parameters by 41.38\% (Fig. \ref{fig1}(e)).

In summary, our main contributions are threefold:
\begin{itemize}
    \item \textbf{First lightweight MoE for MPMP}. We propose Spatiotemporal-Untrammelled Mixture of Experts (ST-MoE), the first framework that integrates spatiotemporal Mamba with dynamic expert routing for multi-person motion prediction. This dual mechanism fundamentally resolves the efficiency-accuracy trade-off.
    \item \textbf{Distinct Spatiotemporal Experts.} We introduce four different spatiotemporal bidirectional mamba experts to flexibly model human motion dynamics, resolving the trammelled spatiotemporal dependency capture in existing multi-person motion prediction methods.
    \item \textbf{Efficient and Low-Parameter.} Extensive experiments on multi-person motion prediction benchmarks demonstrate that our framework attains state-of-the-art performance, while compressing the model size by 41.38\% and achieving 3.6× faster training speed than IAFormer.
\end{itemize}

\section{Related Work}

\begin{figure*}[t]
\centering
\includegraphics[width=0.96\textwidth]{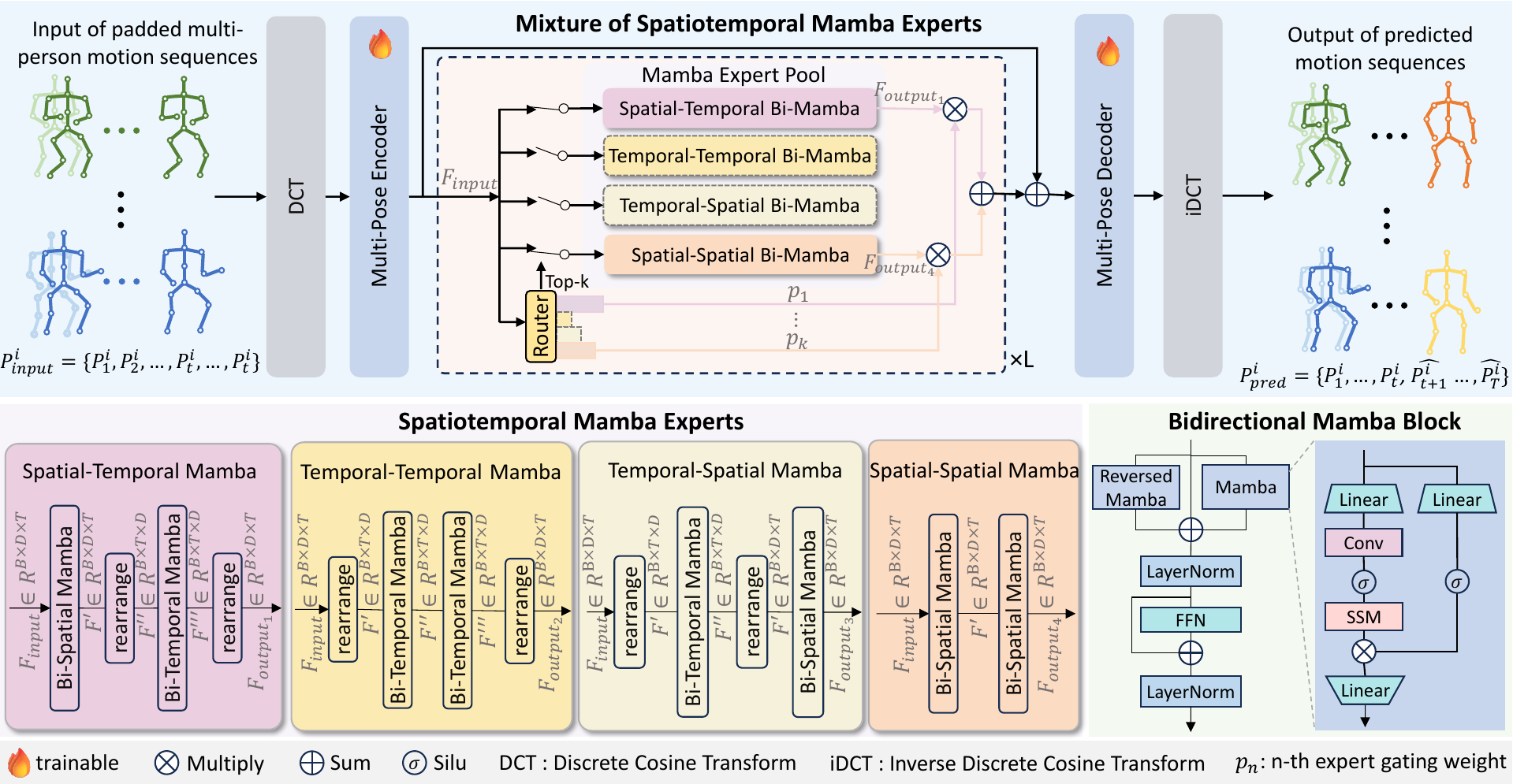} 
\caption{Overview of Spatiotemporal-Untrammelled Mixture of Experts (ST-MoE). 
The padded input motion sequence is first encoded and then adaptively routed by a gating router to distinct spatiotemporal experts. ST-MoE aggregates the outputs of the selected experts and decodes the merged features to predict the future motion sequences.
Each expert in the expert pool consists of pairwise combinations of bidirectional temporal Mamba and bidirectional spatial Mamba, with shared parameters across all experts for both temporal and spatial Mamba components. This design enables comprehensive learning of spatiotemporal dependencies in human motion while maintaining lightweight.}
\label{fig2}
\end{figure*}

\subsubsection{Human Motion Prediction.}
Early research on human motion prediction focused on predicting single-person motion, utilizing RNNs to solve this problem \cite{martinez2017human,fragkiadaki2015recurrent}. However, due to the inherent recursive nature of RNNs leading to problems such as error accumulation, several methods employ Graph Convolutional Networks (GCNs) to explore spatiotemporal dependencies \cite{cui2021towards,ma2022progressively,zhong2022spatio}. Recent research has extended to the more challenging multi-person motion prediction \cite{guo2022multi,wang2021multi,xiao2024multi,xu2023joint,peng2023trajectory}. MRT \cite{wang2021multi} employs spatiotemporal positional encoding to capture constrained spatiotemporal dependencies and IAFormer \cite{xiao2024multi} learns spatiotemporal interactions via attention mechanisms. These methods exhibit high computational complexity and trammelled modeling of complex spatiotemporal dependencies in motion. Conversely, our model extracts diverse spatiotemporal features more efficiently and flexibly.

\subsubsection{Mixture of Experts.}
Mixture of Experts (MoE) was first proposed in  \cite{jacobs1991adaptive} and was later integrated into the Transformer \cite{lepikhin2020gshard,fedus2022switch,du2022glam}. This paradigm replaces Feed Forward Network (FFN) layers with MoE layers, selectively activating only relevant experts per input to achieve adaptive processing, significantly reduce computational cost, while still benefiting from a vast pool of specialized knowledge. More recently, MoE has also seen widespread applications in industrial-scale large language models (LLMs) \cite{liu2024deepseek,teamqwen1,yang2025qwen3}. Unlike previous approaches where each expert employs an identical FFN structure, we utilize distinct spatiotemporal Mamba blocks for individual experts to model human motion more flexibly.

\subsubsection{State Space Model.}
State Space Models (SSMs) have achieved significant success in numerous sequence modeling tasks due to their ability to capture long-range dependencies \cite{gu2023mamba,fu2022hungry,gu2021efficiently,li2024ftmomamba}. Mamba \cite{gu2023mamba} proposes a selective scanning mechanism that dynamically filters irrelevant inputs while maintaining linear inference time and efficient training, establishing itself as a strong Transformer alternative. 
MoE-Mamba \cite{pioro2024moe} combines MoE and Mamba to scale SSMs, matching Transformer performance while retaining computational gains.
MGCN \cite{lin2025mgcn} integrates GCN and Mamba to capture long-term temporal dependencies in traffic flow history data.
However, the  exploration of Mamba’s potential in multi-person motion prediction remains untapped. In this paper, we employ a MoE framework integrating Mamba to capture spatiotemporal dependencies in human motion, significantly boosting model efficiency. 
Unlike MoE-Mamba \cite{pioro2024moe}, our lighter framework integrates Mamba blocks within each expert, while MoE-Mamba alternates Mamba blocks with FFN-based MoE layers.

\section{Method}
\subsection{Overview}
The overview of the proposed ST-MoE is illustrated in Fig. \ref{fig2}. First, the motion sequence is mapped from pose space to feature space via Discrete Cosine Transform (DCT) and a Multi-Pose Encoder, resulting in motion feature \(\textbf{\textit{F}}_{\text{input}}\). Subsequently, \(\textbf{\textit{F}}_{\text{input}}\) is routed to the Mixture of Spatiotemporal Mamba Experts (MoSTME), where each expert consists of a paired combination of bidirectional spatial and temporal Mamba modules in four possible configurations. The router activates specific spatiotemporal experts to flexibly explore the spatiotemporal dependencies in human motion. The outputs \(\textbf{\textit{F}}_{\mathrm{\text{output}}_i}\) from all activated experts are then aggregated, where \(i\) denotes the \(i\)-th activated expert. Finally, the aggregated feature is decoded back from the feature space to the pose space via a Multi-Pose Decoder and Inverse DCT (iDCT), yielding the predicted future motion. 

\subsection{Problem Statement}
Multi-person motion prediction aims to forecast the future joint positions of multiple individuals based on their historical motion. Mathematically, we define the historical motion sequences of the \(i\)-th person  \(\textbf{\textit{P}}_{1: t}^{i}=\left\{\textbf{\textit{P}}_{1}^{i}, \textbf{\textit{P}}_{2}^{i}, \ldots, \textbf{\textit{P}}_{t}^{i}\right\}\in\mathbb{R}^{D \times t}\) and the future motion \(\textbf{\textit{P}}_{t+1:T}^{i}\in\mathbb{R}^{D\times(T-t)}\). The timestamp \(t\) indicates the last observed frame, while \(T\) corresponds to the final predicted frame. The pose dimension \(D\) equals to \(J\times3\), where \(J\) represents the number of joints.
To facilitate the model's learning of future poses, we follow prior work \cite{mao2019learning,dang2021msr,xiao2024multi} by replicating the last observed frame \(\textbf{\textit{P}}_{t}^{i}\) for \(T-t\) times and appending to \(\textbf{\textit{P}}_{1: t}^{i}\), yielding the input sequence \(\textbf{\textit{P}}_{\text{input}}^i = \{ \textbf{\textit{P}}_1^i, \ldots, \textbf{\textit{P}}_t^i, \ldots, \textbf{\textit{P}}_t^i \}\in\mathbb{R}^{D \times T}\).
Our aim is to predict future motion \(\textbf{\textit{P}}_{\text{pred}}^i = \{ \textbf{\textit{P}}_1^i, \ldots, \textbf{\textit{P}}_{t}^i , \hat{\textbf{\textit{P}}}_{t+1}^i, \ldots, \hat{\textbf{\textit{P}}}_{T}^i \}\in\mathbb{R}^{D \times T}\) from padded motion sequences \(\textbf{\textit{P}}_{\text{input}}^i\).

\subsection{Multi-Pose Encoder and Decoder}
GCNs have demonstrated superior capability in capturing spatiotemporal dependencies among joints in human motion, and are therefore widely adopted in motion prediction tasks \cite{mao2019learning}. For example, IAFormer \cite{xiao2024multi} employs GCN-based encoder and decoder modules to extract features from multi-person motion. Inspired by this design, we adopt the multi-pose encoder/decoder architecture from IAFormer to enhance the model's capacity in representing complex pose dynamics.
In addition, to further improve representation compactness and effectively capture the smooth dynamics of human motion, we apply the DCT before the encoder and iDCT after the decoder \cite{mao2019learning,wang2021multi}. The encoding process is as follows:
\begin{equation}
    \textbf{\textit{F}}_{\text{input}}^{i} = \texttt{ME}(\texttt{DCT}(\textbf{\textit{P}}_{\text{input}}^{i})),
    \label{eq:1}
\end{equation}
where \(\texttt{ME}(\cdot)\) represents Multi-Pose Encoder and \(\texttt{DCT}(\cdot)\) represents Discrete Cosine Transform. The \(\textbf{\textit{F}}_{\text{input}}^{i}\) denotes the motion feature in the feature space of the \(i\)-th person. 
\subsection{Mixture of Spatiotemporal Mamba Experts}
As illustrated in Fig. \ref{fig1}(a)-(c), existing approaches often employ fixed spatio-temporal positional encodings or spatio-temporal attention mechanisms to capture spatio-temporal features, resulting in trammelled patterns. Inspired by the flexible modeling capability of MoE \cite{shazeer2017outrageously,yun2024flex}, we introduce the Mixture of Spatiotemporal Mamba Experts to address the challenges mentioned above. First, the encoded features \(\textbf{\textit{F}}_{\text{input}}\in \mathbb{R}^{B \times D \times T}\) are input into the Mamba Expert Pool and the router, where \(B\) denotes batch size. Furthermore, the router implements sparse activation by dynamically selecting only the top-k experts through an MLP-based gating network. It then calculates weights specifically for these activated experts and aggregates their outputs via weighted summation. This entire process can be expressed by the following formula:
\begin{equation}
\begin{aligned}
\textbf{\textit{E}}_{\text{output}} &= \sum_{e=1}^{N} 
\textbf{\textit{f}}_e(\textbf{\textit{F}}_{\text{input}})
\textbf{\textit{p}}_e, \\
\textbf{\textit{p}}_e &= \texttt{softmax}(\texttt{TopK}(g(\textbf{\textit{F}}_{\text{input}}), k))_e,
\end{aligned}
\label{eq:2}
\end{equation}
where \(N\) denotes the total number of experts, \(\textbf{\textit{p}}_e\) represents the dynamic weight of the \(e\)-th expert, \(\textbf{\textit{f}}_e(\cdot)\) corresponds to the \(e\)-th expert, and \(g(\cdot)\) denotes gating function. The \(\texttt{TopK}(\cdot, k)\) preserves only the original values of the top-k entries in the vector while setting all other entries to \(-\infty\). After the softmax operation, these entries assigned \(-\infty\) effectively approximate zero, achieving sparse activation. To enable holistic learning of spatiotemporal motion features, we choose to activate all experts, which is empirically validated by our later experimental results.

\subsection{Bidirectional Spatiotemporal Mamba} 
Previous approaches incorporate temporal or spatial information as biases into self-attention mechanisms to extract spatio-temporal features \cite{peng2023trajectory,xu2023joint,xiao2024multi}. However, due to the quadratic complexity of self-attention, these approaches suffer from high computational costs in spatiotemporal modeling. To address this issue, we introduce Mamba with linear complexity into each expert module to reduce computational overhead. Specifically, we construct four structurally distinct experts by combining bidirectional temporal Mamba and bidirectional spatial Mamba in different configurations. Unlike traditional MoE designs that employ a single feed-forward network (FFN) architecture shared by all experts, this design allows each expert to specialize in capturing distinct types of spatiotemporal features, thereby significantly enhancing the model’s flexibility and expressiveness in modeling complex spatiotemporal dynamics.

As illustrated in Fig. \ref{fig2} , we devise four distinct types of spatiotemporal Mamba experts. Taking the Spatial-Temporal Mamba expert as an example, the input features \(\textbf{\textit{F}}_{\text{input}}\) first pass through a bidirectional spatial Mamba to extract spatial features, followed by temporal processing. In this way, Spatial-Temporal Mamba can be formulated by:
\begin{equation}
\begin{aligned}
    \textbf{\textit{F}}'' &= \texttt{rearrange}(\texttt{Bi-SMamba}(\textbf{\textit{F}}_{\text{input}})),  \\
    \textbf{\textit{F}}_{\text{output}_1} &= \texttt{rearrange}(\texttt{Bi-TMamba}(\textbf{\textit{F}}'')),
\end{aligned}
\label{eq:3}
\end{equation}
where \(\textbf{\textit{F}}_{\text{input}} \in \mathbb{R}^{B \times D \times T}\) denotes the input feature, \(\texttt{Bi-SMamba}(\cdot)\) and \(\texttt{Bi-TMamba}(\cdot)\) denotes bidirectional spatial Mamba and bidirectional temporal Mamba, respectively. \(\texttt{rearrange}(\cdot)\) represents the tensor transposition operation. The \(\textbf{\textit{F}}'' \in \mathbb{R}^{B \times T \times D}\) denotes the feature after transposition and \(\textbf{\textit{F}}_{\mathrm{output}_1} \in \mathbb{R}^{B \times D \times T}\) denotes the final output. Other experts adopt similar operations, differing only in the order of data processing. To facilitate the learning of spatio-temporal features, each expert shares bidirectional temporal Mamba and bidirectional spatial Mamba, which can further reduce model parameters.

Moreover, the unidirectional modeling nature of the original Mamba limits its ability to capture global dependencies. To address this issue, we introduce a bidirectional scanning mechanism into both spatial and temporal Mamba modules to enhance global dependency modeling in complex motion sequences. First, the spatial Mamba performs bidirectional scanning along the spatial dimension of input features to capture spatial dependencies, then optimizes learning through residual connections. Similarly, the temporal Mamba operates along the temporal dimension. It can be formulated as follows:
\begin{equation}
\begin{aligned}
     {\textbf{\textit{f}}}_o^{s} &= \texttt{SMamba}(\overrightarrow{{\textbf{\textit{f}}}_s}) + \texttt{SMamba}(\overleftarrow{{\textbf{\textit{f}}}_s}) +
    \overrightarrow{{\textbf{\textit{f}}}_s}, \\
    {\textbf{\textit{f}}}_o^{t} &= \texttt{TMamba}(\overrightarrow{{\textbf{\textit{f}}}_t}) + \texttt{TMamba}(\overleftarrow{{\textbf{\textit{f}}}_t}) +
    \overrightarrow{{\textbf{\textit{f}}}_t}, 
\end{aligned}
\label{eq:4}
\end{equation}
where \(\overrightarrow{{\textbf{\textit{f}}}_s} \in \mathbb{R}^{B \times D \times T}\) and \(\overrightarrow{{\textbf{\textit{f}}}_t} \in \mathbb{R}^{B \times T \times D}\) respectively represent the input features of the spatial Mamba and temporal Mamba. \(\overleftarrow{{\textbf{\textit{f}}}_s}\) and \(\overleftarrow{{\textbf{\textit{f}}}_t}\) denote spatially-reversed feature and temporally-reversed feature, respectively. \(\texttt{SMamba}(\cdot)\) and \(\texttt{TMamba}(\cdot)\) denote spatial Mamba and temporal Mamba, respectively. \({\textbf{\textit{f}}}_o^{s}\) and \({\textbf{\textit{f}}}_o^{t}\) denotes the output of bidirectional spatial Mamba and bidirectional temporal Mamba, respectively. Second, we employ Layer Normalization to stabilize the training process and leverage an FFN to enhance feature representation capabilities. Subsequently, residual connections are utilized to facilitate model learning:
\begin{equation}
\begin{aligned}
    {\textbf{\textit{F}}}_{o}^\star &= \texttt{LN}(\texttt{LN}({\textbf{\textit{f}}}_o^\star) + \texttt{FFN}(\texttt{LN}({\textbf{\textit{f}}}_o^\star))), \\
\end{aligned}
\label{eq:5}
\end{equation}
where the superscript is \(\star \in \{s, t\}\), \({\textbf{\textit{F}}}_{o}^\star\) is the output of bidirectional spatiotemporal Mamba, \(\texttt{LN}(\cdot)\) is Layer Normalization, and \(\texttt{FFN}(\cdot)\) is Feed Forward Network. Then, \({\textbf{\textit{F}}}_{o}^s\) and \({\textbf{\textit{F}}}_{o}^t\) undergo distinct propagation orders across spatiotemporal experts, 
yielding expert-specific outputs $\{ \textbf{\textit{F}}_{\text{output}_i} \}_{i=1}^{4}$. Afterwards, we aggregate the outputs of activated experts to generate the \(l\)-th layer representation \(\textbf{\textit{E}}_{\text{output}}^{l}\)  as input to the next MoE layer. Finally, we apply residual connections and decode the final features \(\textbf{\textit{E}}_{\text{output}}^{L}\) to predict future motion:
\begin{equation}
    \textbf{\textit{P}}_{\text{pred}}^i = \texttt{iDCT}(\texttt{MD}(\textbf{\textit{F}}_{\text{input}}^{i} + \textbf{\textit{E}}_{\text{output}}^{L})),
    \label{eq:6}
\end{equation}
where \(\textbf{\textit{P}}_{\text{pred}}^i\) denotes the final predicted motion of the \(i\)-th person, \(\texttt{iDCT}(\cdot)\) denotes Inverse Discrete Cosine Transform and \(\texttt{MD}(\cdot)\) represents Multi-Pose Decoder.

\subsection{Loss Function}
To constrain history and future joint positions in human motion, we employ spatial loss \(L_s\), which is calculated as:
\begin{equation}
\begin{aligned}
L_s &=  \frac{\lambda}{J \cdot M \cdot t} \sum_{m=1}^{M} \sum_{j=1}^{J} \sum_{i=1}^{t} \left\| \hat{\textbf{\textit{P}}}_{i,j}^m - \textbf{\textit{P}}_{i,j}^m \right\|^2  \\
&\quad + \frac{1}{J \cdot M \cdot (T - t)} \sum_{m=1}^{M} \sum_{j=1}^{J} \sum_{i=t+1}^{T} \left\| \hat{\textbf{\textit{P}}}_{i,j}^m - \textbf{\textit{P}}_{i,j}^m \right\|^2, 
\label{eq:ls}
\end{aligned}
\end{equation}
where \(\hat{\textbf{\textit{P}}}_{i,j}^m\) and \(\textbf{\textit{P}}_{i,j}^m\) represent the predicted and ground-truth position of the \(j\)-th joint for the \(m\)-th person at timestamp \(i\). \(M\) is the number of humans and \(\lambda\) is the weight coefficient.

To mitigate temporal jitter in predicted movements, following IAFormer \cite{xiao2024multi}, we employ temporal consistency loss \(L_t\), which is formulated as:
\begin{equation}
    L_t = \texttt{MSE}(\texttt{Conv}(\textbf{\textit{P}}_{\text{pred}}), \texttt{Conv}(\textbf{\textit{P}}_{\text{gt}})),
    \label{eq:lt}
\end{equation}
where \(\texttt{Conv}(\cdot)\) is Convolutional Neural Networks for feature mapping. \(\texttt{MSE}(\cdot)\) represents Mean Squared Error.

Finally, we perform end-to-end training by optimizing the aggregated final loss function, defined as:
\begin{equation}
    L = \alpha L_s+\beta L_t,
    \label{eq:loss}
\end{equation}
where \(\alpha\) and \(\beta\) are the weight coefficient.

\section{Experiments}
\subsection{Datasets}
Following IAFormer \cite{xiao2024multi}, to verify the effectiveness of ST-MoE, we conduct experiments on four multi-person motion datasets: CMU-Mocap \cite{CMUGraphicsLabMocap2003}, UMPM \cite{van2011umpm}, Mix1 and Mix2 \cite{peng2023trajectory}, CHI3D \cite{chi3d}.
\subsubsection{CMU-Mocap (UMPM).}
CMU-Mocap (UMPM) is a synthetic three-person motion dataset created by integrating UMPM into CMU-Mocap. The training set contains 13,000 sequences, the test set contains 3,000 sequences, and each sequence comprises 75 frames.

\subsubsection{Mix1 and Mix2.} 
To validate the model's generalization capability, we train on the CMU-Mocap (UMPM) dataset and evaluate on the Mix1 and Mix2 datasets. Mix1 contains 6 individuals while Mix2 comprises 10 individuals. Both datasets are constructed by MuPoTS-3D \cite{mehta2018single}, 3DPW \cite{von2018recovering}, and test data from CMU-Mocap(UMPM). Each dataset consists of 1,000 motion sequences, with each sequence containing 75 frames.

\subsubsection{CHI3D.} 
Unlike artificially mixed datasets, CHI3D is a lab-based accurate 3d motion capture dataset containing two individuals. It better reflects real-world scenarios, making it particularly suitable for capturing complex spatiotemporal dependencies in multi-person motions.

\subsection{Metrics}
Following IAFormer \cite{xiao2024multi}, we adopt the mean per Joint Position Error (JPE) and Aligned Position Error (APE) as evaluation metrics. JPE measures global joint position errors, including overall body displacement. APE aligns the root joint to remove global movement, focusing on evaluating pose-specific errors. \textit{More details in appendix. }

\begin{table*}[t]
\centering
\small 
\setlength{\tabcolsep}{1mm} 
\begin{tabular}{cl|cccc|cccc|cccc}
\hline
\multicolumn{2}{c|}{\multirow{2}{*}{Method}} & \multicolumn{4}{c|}{CMU-Mocap (UMPM) 3 persons} & \multicolumn{4}{c|}{Mix1 6 persons} & \multicolumn{4}{c}{Mix2 10 persons} \\ 
\cline{3-14} 
\multicolumn{2}{c|}{} & 0.2s↓ & 0.6s↓ & 1.0s↓ & Avg↓ & 0.2s↓ & 0.6s↓ & 1.0s↓ & Avg↓ & 0.2s↓ & 0.6s↓ & 1.0s↓ & Avg↓ \\ 
\hline
\multicolumn{1}{c|}{\multirow{7}{*}{JPE}} & MSR-GCN \cite{dang2021msr} & 53 & 146 & 231 & 143 & 49 & 132 & 220 & 134 & 60 & 153 & 243 & 152 \\
\multicolumn{1}{c|}{} & HRI \cite{mao2020history} & 49 & 130 & 207 & 129 & 51 & 141 & 233 & 142 & 52 & 140 & 224 & 139 \\
\multicolumn{1}{c|}{} & MRT* \cite{wang2021multi} & 36 & 115 & 193 & 114 & 37 & 122 & 212 & 124 & 38 & 126 & 214 & 126 \\
\multicolumn{1}{c|}{} & TBIFormer* \cite{peng2023trajectory} & \textbf{30} & 109 & 182 & 107 & 34 & 121 & 209 & 121 & 34 & 118 & 198 & 117 \\
\multicolumn{1}{c|}{} & JRFormer* \cite{xu2023joint} & 32 & 104 & 161 & 99 & \textbf{32} & {\underline{109}} & \textbf{184} & \textbf{108} & 36 & 125 & 211 & 124 \\
\multicolumn{1}{c|}{} & T2P* \cite{jeong2024multi} & 38 & 102 & 158 & 99 & - & - & - & - & - & - & - & - \\
\multicolumn{1}{c|}{} & IAFormer* \cite{xiao2024multi} & 32 & {\underline{96}} & {\underline{159}} & 96 & 36 & 112 & 193 & 114 & {\underline{36}} & {\underline{108}} & {\underline{181}} & {\underline{108}} \\
\multicolumn{1}{c|}{} & \textbf{ST-MoE* (Ours)} & {\underline{31}} & \textbf{95} & \textbf{158} & \textbf{95} & {\underline{34}} & \textbf{108} & {\underline{187}} & {\underline{110}} & \textbf{34} & \textbf{106} & \textbf{179} & \textbf{107} \\
\hline
\multicolumn{1}{c|}{\multirow{7}{*}{APE}} & MSR-GCN \cite{dang2021msr} & 46 & 106 & 137 & 96 & 41 & 92 & 120 & 84 & 48 & 110 & 148 & 102 \\
\multicolumn{1}{c|}{} & HRI \cite{mao2020history} & 41 & 97 & 130 & 89 & 38 & 92 & 122 & 84 & 41 & 100 & 133 & 91 \\
\multicolumn{1}{c|}{} & MRT* \cite{wang2021multi} & 36 & 108 & 159 & 101 & 36 & 109 & 166 & 104 & 38 & 115 & 178 & 110 \\
\multicolumn{1}{c|}{} & TBIFormer* \cite{peng2023trajectory} & 27 & 84 & 118 & 76 & 28 & 81 & 113 & 74 & 30 & 89 & 124 & 81 \\
\multicolumn{1}{c|}{} & JRFormer* \cite{xu2023joint} & \textbf{20} & 78 & 114 & 71 & \textbf{21} & 73 & 105 & 66 & \textbf{22} & 82 & 120 & {\underline{75}} \\
\multicolumn{1}{c|}{} & T2P* \cite{jeong2024multi} & 34 & 84 & 116 & 78 & - & - & - & - & - & - & - & - \\
\multicolumn{1}{c|}{} & IAFormer* \cite{xiao2024multi} & 23 & {\underline{71}} & \textbf{103} & {\underline{66}} & 23 & {\underline{71}} & {\underline{101}} & {\underline{65}} & 24 & {\underline{76}} & \textbf{108} & \textbf{69} \\
\multicolumn{1}{c|}{} & \textbf{ST-MoE* (Ours)} & {\underline{22}} & \textbf{70} & {\underline{104}} & \textbf{65} & {\underline{22}} & \textbf{69} & \textbf{100} & \textbf{64} & {\underline{23}} & \textbf{75} & {\underline{109}} & \textbf{69} \\
\hline
\end{tabular}
\caption{Performance comparison (in mm) on mixed multi-person datasets. * means multi-person motion prediction method. The best results are in \textbf{bold} and the second-best ones are \underline{underlined}.}
\label{table1}
\end{table*}

\subsection{Implementation Details}
\textbf{Network Architecture.} The Multi-Pose Encoder/Decoder employs a 3-layer GCN structure, while the MoE module uses a single-layer design with a one-layer MLP gating function. Observed sequences are set to 50 frames (2s) to predict future 25 frames (1s). The pose dimension \(D\) is set to 45. \textbf{Reproducibility.} We train our model with a batch size of 96 and adopt the Adam \cite{kingma2014adam} optimizer with an initial learning rate of 0.01. The learning rate is decayed exponentially by a factor of \(0.1^{1/50}\) per epoch. 
The loss weighting coefficients are configured with \(\alpha=1\), \(\beta=1\), \(\lambda=0.1\). 
Training is performed on one RTX 3090 GPU.

\subsection{Comparison with SOTA Methods}
\subsubsection{Results on CMU-Mocap (UMPM).}
As shown in Table \ref{table1}, our method ST-MoE achieves SOTA results on CMU-Mocap (UMPM), outperforming JRFormer \cite{xu2023joint} by 4 mm JPE and 6 mm APE in average metrics. 
This improvement stems from ST-MoE's ability to adaptively capture complex spatiotemporal dependencies through four distinct experts, whereas existing methods exhibit suboptimal performance due to their constrained capacity in modeling diverse spatiotemporal patterns.

\subsubsection{Results on Mix1 and Mix2.}
On Mix1 and Mix2 datasets, ST-MoE demonstrates strong generalization as the number of individuals increases. It consistently outperforms baselines: on Mix1, gains are 4 mm JPE and 1 mm APE over IAFormer \cite{xiao2024multi}; on Mix2, gains are 17 mm JPE and 6 mm APE over JRFormer. 
This demonstrates superior capability of ST-MoE to capture more complex spatiotemporal dependencies in scenarios with more individuals.

\subsubsection{Results on CHI3D.}
Compared to manually mixed datasets, CHI3D reflects spatio-temporal dependencies in multi-person scenarios more authentically \cite{xiao2024multi}. As shown in Table \ref{table2}, ST-MoE achieves state-of-the-art performance on CHI3D, exhibiting 8 mm lower average JPE than IAFormer and 21 mm lower than TBIFormer \cite{peng2023trajectory}.
These results demonstrate ST-MoE's powerful capability to capture spatiotemporal dependencies in real-world scenarios.

\subsubsection{Model Efficiency and Parameter Economy.}
As shown in Fig. \ref{fig1}(e), ST-MoE surpasses IAFormer in performance while achieving 3.6x faster training and 41.38\% fewer parameters, attributed to Mamba's linear time complexity, avoiding the high computational cost of attention mechanisms. For fair comparison, we train all models with a batch size of 96. \textit{More results in Appendix.}

\begin{table}[t]
\centering
\small 
\setlength{\tabcolsep}{0.5mm} 
\begin{tabular}{cl|c|c|c|c|c|c}
\hline
\multicolumn{2}{c|}{Method}                                         & 0.2s↓       & \multicolumn{1}{l|}{0.4s↓} & 0.6s↓        & \multicolumn{1}{l|}{0.8s↓} & 1.0s↓        & Avg↓         \\ \hline
\multicolumn{1}{c|}{\multirow{4}{*}{JPE}} & PGBIG \cite{ma2022progressively}                   & 69          & 130                        & 181          & 223                        & 258          & 172          \\
\multicolumn{1}{c|}{}                     & TBIFormer*              & 45          & 95                         & 145          & 192                        & 233          & 142          \\
\multicolumn{1}{c|}{}                     & IAFormer*               & \textbf{39} & {\underline {83}}                   & {\underline {129}}    & {\underline {176}}                  & {\underline {218}}    & {\underline {129}}    \\
\multicolumn{1}{c|}{}                     & \textbf{ST-MoE* (Ours)} & {\underline {44}}    & \textbf{79}                & \textbf{123} & \textbf{161}               & \textbf{200} & \textbf{121} \\ \hline
\end{tabular}%
\caption{JPE results (in mm) on CHI3D dataset. }
\label{table2}
\end{table}

\begin{table}[t]
\centering
\setlength{\tabcolsep}{1mm} 
\small 
\begin{tabular}{c|cccc|cccc}
\hline
\multirow{2}{*}{Method} & \multicolumn{4}{c|}{JPE}                                       & \multicolumn{4}{c}{APE}                                        \\ \cline{2-9} 
                        & 0.2s↓         & 0.6s↓         & 1.0s↓          & Avg↓          & 0.2s↓         & 0.6s↓         & 1.0s↓          & Avg↓          \\ \hline
Baseline                & 35.7          & 113.2         & 184.3          & 111.1         & 24.7          & 81.1          & 114.1          & 73.3          \\ \hline
+ST                     & 37.1          & 105.8         & 170.7          & 104.5         & 26.0          & 77.3          & 108.9          & 70.7          \\
+TT                     & 33.2          & 99.1          & 162.1          & 98.1          & 23.0          & 72.1          & 104.1          & 66.4          \\
+TS                     & 34.7          & 100.8         & 164.8          & 100.1         & 24.6          & 74.7          & 106.9          & 68.7          \\
+SS                     & 33.6          & 99.1          & 162.4          & 98.3          & 23.9          & 74.9          & 105.8          & 68.2          \\ \hline
+All                    & \textbf{31.4} & \textbf{95.3} & \textbf{158.3} & \textbf{95.0} & \textbf{22.1} & \textbf{70.4} & \textbf{103.8} & \textbf{65.4} \\ \hline
\end{tabular}%
\caption{Ablation study for effectiveness of distinct experts on CMU-Mocap (UMPM) dataset.}
\label{table3}
\end{table}

\begin{figure}[t]
\centering
\includegraphics[width=0.465\textwidth]{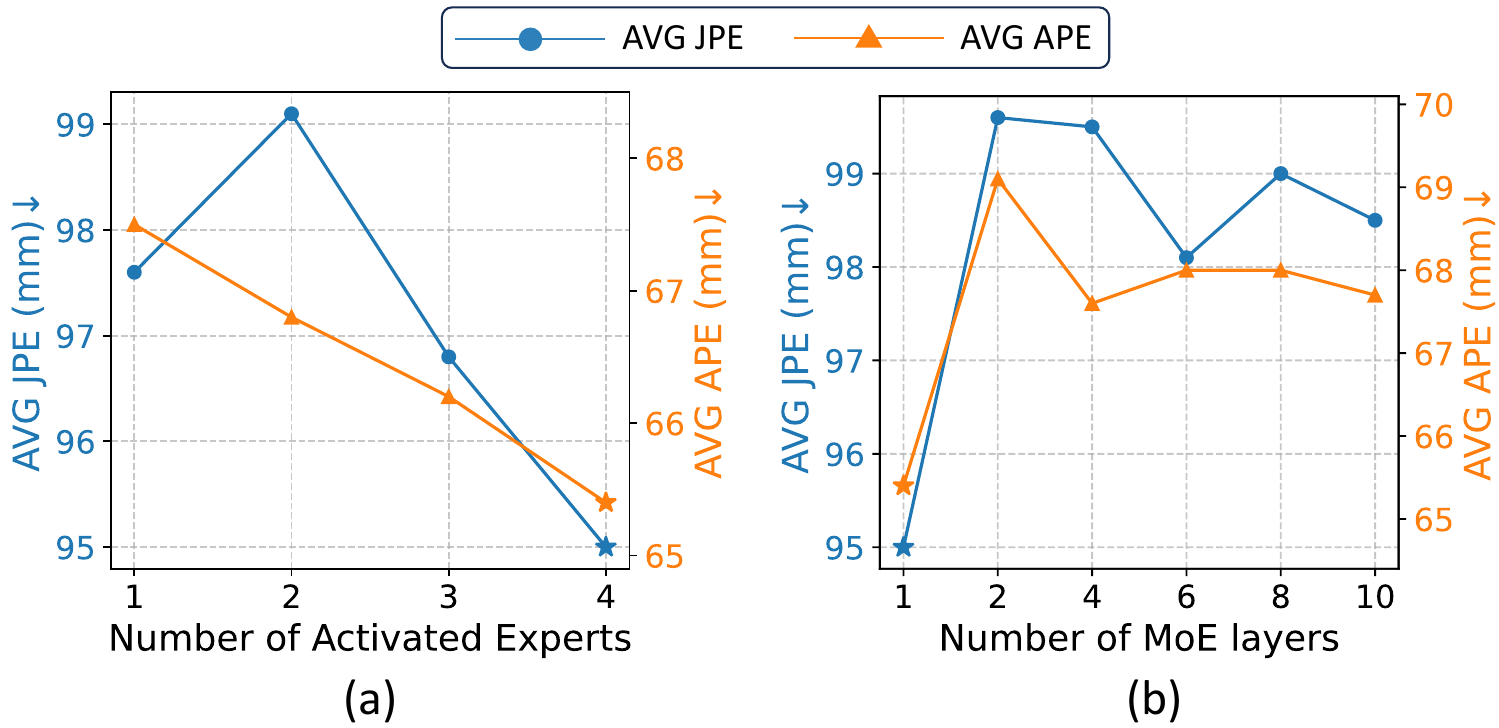} 
\caption{(a): Impact of different numbers of activated experts. (b): Impact of different numbers of MoE layers. Both are tested on the CMU-Mocap (UMPM) dataset.}
\label{fig3}
\end{figure}

\begin{figure}[t]
\centering
\includegraphics[width=0.48\textwidth]{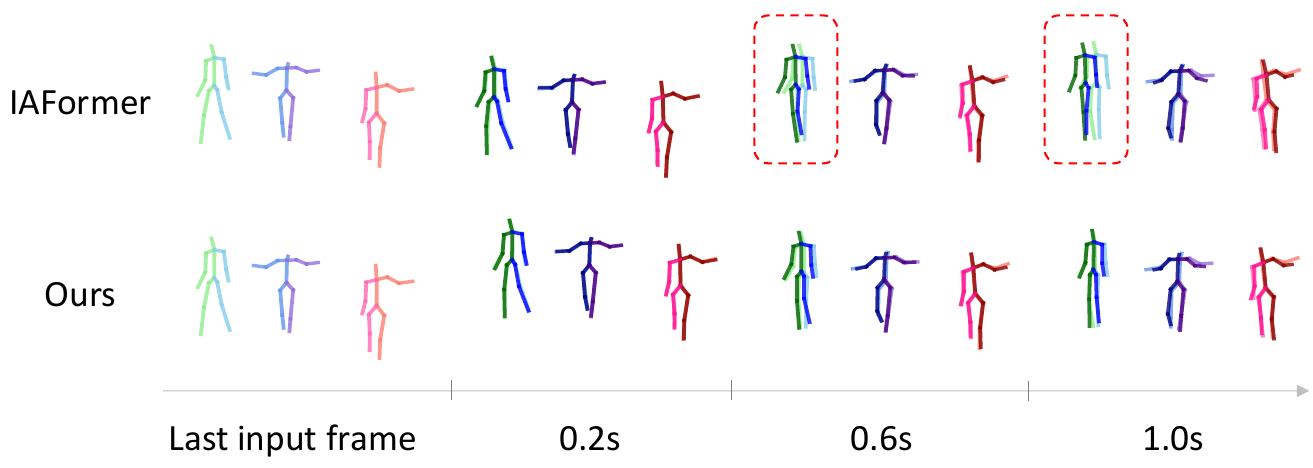} 
\caption{Visualization comparison on CMU-Mocap (UMPM) dataset. Dark-colored lines represent predicted motion, while light-colored lines indicate ground truth.}
\label{visualization}
\end{figure}

\subsection{Ablation Studies}
\begin{figure}[t]
\centering
\includegraphics[width=0.42\textwidth]{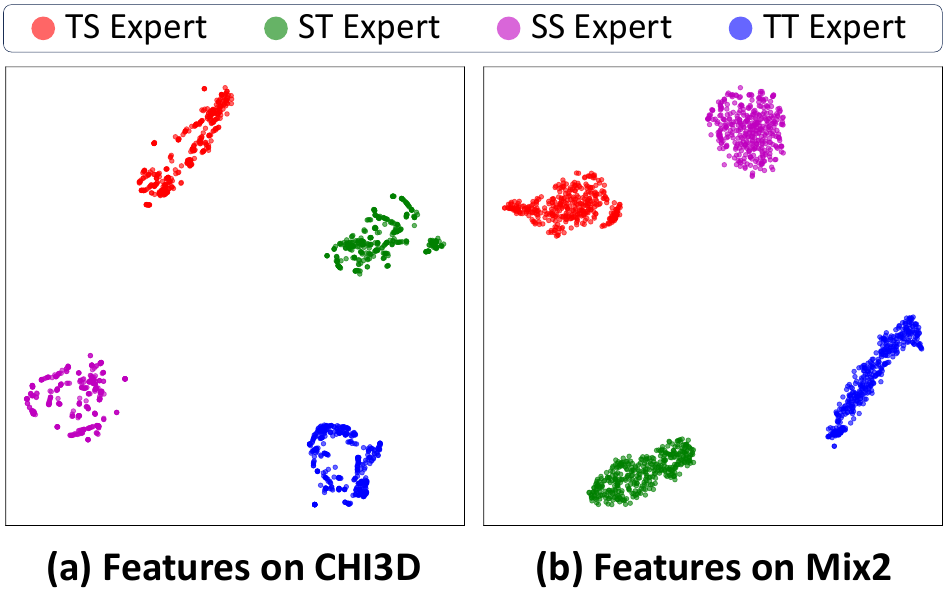} 
\caption{The t-SNE visualization of features learned by four distinct experts on CHI3D and Mix2 dataset. 300 samples are randomly selected. Best view in color.}
\label{tsne}
\end{figure}

\subsubsection{Effectiveness of Distinct Experts.}
To validate the effectiveness of collaborative expert interactions, we systematically replaced the four heterogeneous spatiotemporal experts with each individual type. The number of experts is set to 4. The baseline model incorporates solely Multi-Pose Encoder/Decoder, while ST, TT, TS, and SS correspond to the four expert models illustrated in Fig. \ref{fig2}. In Table \ref{table3}, we can observe that: i) Integrating any single expert type enhances performance over the baseline, exemplified by TT expert reducing average JPE by 13.0mm and average APE by 6.9mm; ii) Combining heterogeneous experts significantly outperforms models using uniform expert types, demonstrating the critical value of specialized collaborative modeling.

\subsubsection{Impact of Different Numbers of Activated Experts.}
To investigate the optimal number of activated experts for enhanced performance, we progressively activate the four distinct spatiotemporal experts by increasing the parameter \(k\) in Eq. (\ref{eq:2}) from 1 to 4. As illustrated in Fig. \ref{fig3}(a), the average APE and JPE demonstrate an overall decreasing trend as more experts are activated. Optimal performance is achieved when all four experts are fully activated. 
This indicates that as the number of activated experts increases, spatio-temporal features can be more flexibly processed by different experts, thereby enhancing prediction performance.

\subsubsection{Impact of Varying Numbers of MoE Layers.}
As shown in Fig. \ref{fig3}(b), we investigate the impact of varying numbers of MoE layers on predictive performance using the CMU-Mocap (UMPM) dataset. We observe that employing a single MoE layer yields the most effective results, as stacking more layers may lead to overfitting and training becomes more difficult. \textit{More results in Appendix.}

\begin{figure}[t]
\centering
\includegraphics[width=0.47\textwidth]{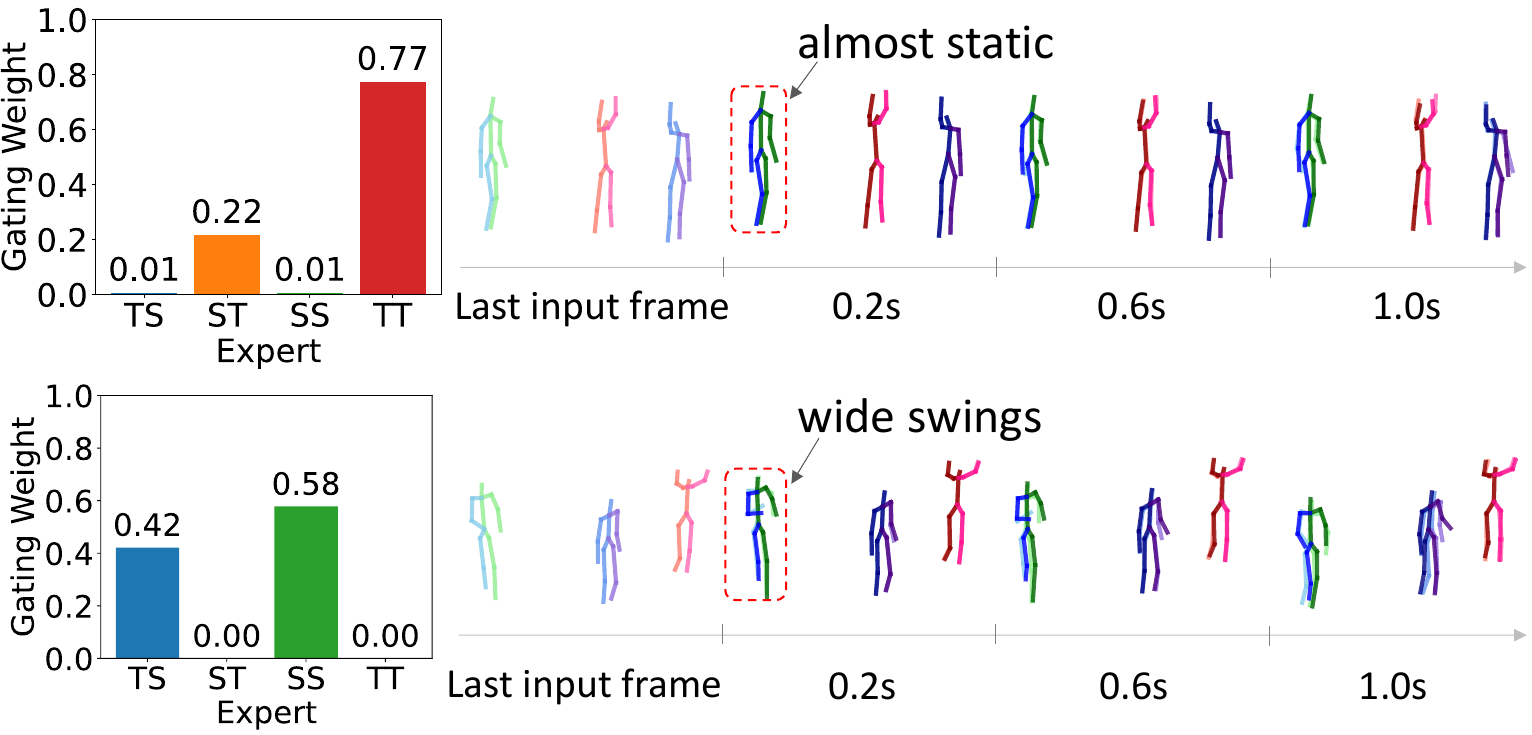} 
\caption{Visualization of adaptive gating  weights of four experts and corresponding predicted motion on CMU-Mocap (UMPM). Distinct experts capture different spatiotemporal patterns with adaptive weights.}
\label{ada_expert}
\end{figure}

\subsection{Qualitative Analysis}
\subsubsection{Visualization of Prediction Results.}
Fig. \ref{visualization} presents a comparative visualization of ST-MoE against IAFormer \cite{xiao2024multi} on the CMU-Mocap (UMPM) dataset. The left column depicts the last input frames, while the three right columns show predictions at different timestamps. The individual on the left begins moving leftward before abruptly stopping. Due to insufficient modeling of spatiotemporal dependencies, IAFormer fails to capture this ``dynamic-to-static'' motion pattern, resulting in prediction drift. In contrast, our approach captures this transition by leveraging four specialized spatiotemporal experts, achieving higher accuracy in motion forecasting. \textit{More visualization results in Appendix.}

\subsubsection{The t-SNE of Features Learned by Four Experts.} 
As shown in Fig. \ref{tsne}, we visualize the feature spaces of the four distinct spatiotemporal experts via t-SNE on both CHI3D and Mix2 datasets. The former embodies more authentic spatiotemporal dependencies, while the latter contains a larger number of individuals exhibiting more complex spatiotemporal patterns. With 300 randomly selected samples processed by all experts, we observe that each expert forms well-separated clusters with pronounced inter-cluster distinctions. This empirically verifies that the four experts capture divergent spatiotemporal motion patterns, while our ST-MoE framework adaptively integrates these representations to achieve enhanced prediction precision.

\subsubsection{Flexible Spatiotemporal Modeling.} 
To investigate how the four experts collaborate, we visualize the adaptive gating weights corresponding to each expert and predicted motions on the CMU-Mocap (UMPM) dataset. As illustrated in Fig. \ref{ada_expert}, adaptive selection of TT/ST experts yields almost static motions, reflecting their focus on spatiotemporal patterns with limited spatial variation. Conversely, when SS/TS experts are activated, the corresponding motion visualization exhibits wide arm swings during running, demonstrating their strength in modeling spatially dynamic patterns. ST-MoE dynamically allocates weights to distinct spatiotemporal experts via adaptive gating, enabling more flexible capture of spatiotemporal dependencies in multi-person motion.

\section{Conclusion and Limitations}
\subsubsection{Conclusion.}
In this work, we propose the first lightweight framework \textbf{ST-MoE} for efficient multi-person motion prediction. 
To overcome trammelled spatiotemporal modeling, we introduce four specialized experts that capture distinct spatiotemporal patterns, enabling comprehensive and flexible learning of motion pattern through adaptive expert selection.
To resolve the high computational complexity of existing methods, we design bidirectional spatiotemporal Mamba blocks as the backbone network for each expert, significantly reducing computational overhead. Exhaustive experiments validate that our approach achieves an optimal trade-off between efficiency and performance.

\subsubsection{Limitations and Future Work.}
While our approach achieves a balance between efficiency and performance, it is limited to deterministic motion. Future work will targets stochastic multi-person motion prediction, exploring MoE for modeling diversified motion distributions.

\section{Acknowledgments}
The work is supported by the National Natural Science Foundation of China (Grant No. U25A20442, 62222207, 62427808, 62472208).

\bibliography{references}

@String(PAMI = {IEEE Trans. Pattern Anal. Mach. Intell.})

@String(CVPR= {IEEE Conf. Comput. Vis. Pattern Recog.})

@String(ICCV= {Int. Conf. Comput. Vis.})

@String(ECCV= {Eur. Conf. Comput. Vis.})

@String(NIPS= {Adv. Neural Inform. Process. Syst.})

@String(TIP  = {IEEE Trans. Image Process.})

@String(TMM  = {IEEE Trans. Multimedia})

@String(AAAI = {AAAI})

@String(PAMI  = {IEEE TPAMI})

@String(CVPR  = {CVPR})

@String(ICCV  = {ICCV})

@String(ECCV  = {ECCV})

@String(NIPS  = {NeurIPS})

@String(TIP   = {IEEE TIP})

@String(TMM   =	{IEEE TMM})

@string(ICML={ICML})

@inproceedings{van2011umpm,
  title={Umpm benchmark: A multi-person dataset with synchronized video and motion capture data for evaluation of articulated human motion and interaction},
  author={Van der Aa, NP and Luo, Xinghan and Giezeman, Geert-Jan and Tan, Robby T and Veltkamp, Remco C},
  booktitle={ICCV Workshops},
  pages={1264--1269},
  year={2011}
}

@misc{CMUGraphicsLabMocap2003,
  title = {Cmu graphics lab motion capture database},
  author = {CMU-Graphics-Lab},
  howpublished = {\url{http://mocap.cs.cmu.edu/}},
  year = {2003}
}

@inproceedings{xu2023joint,
  title={Joint-relation transformer for multi-person motion prediction},
  author={Xu, Qingyao and Mao, Weibo and Gong, Jingze and Xu, Chenxin and Chen, Siheng and Xie, Weidi and Zhang, Ya and Wang, Yanfeng},
  booktitle=CVPR,
  pages={9816--9826},
  year={2023}
}

@inproceedings{xiao2024multi,
  title={Multi-person Pose Forecasting with Individual Interaction Perceptron and Prior Learning},
  author={Xiao, Peng and Xie, Yi and Xu, Xuemiao and Chen, Weihong and Zhang, Huaidong},
  booktitle= ECCV,
  pages={402--419},
  year={2024},
}

@article{wang2021multi,
  title={Multi-person 3d motion prediction with multi-range transformers},
  author={Wang, Jiashun and Xu, Huazhe and Narasimhan, Medhini and Wang, Xiaolong},
  journal=NIPS,
  volume={34},
  pages={6036--6049},
  year={2021}
}

@inproceedings{peng2023trajectory,
  title={Trajectory-aware body interaction transformer for multi-person pose forecasting},
  author={Peng, Xiaogang and Mao, Siyuan and Wu, Zizhao},
  booktitle=CVPR,
  pages={17121--17130},
  year={2023}
}

@article{gu2023mamba,
  title={Mamba: Linear-time sequence modeling with selective state spaces},
  author={Gu, Albert and Dao, Tri},
  journal={arXiv preprint arXiv:2312.00752},
  year={2023}
}

@inproceedings{butepage2017deep,
  title={Deep representation learning for human motion prediction and classification},
  author={Butepage, Judith and Black, Michael J and Kragic, Danica and Kjellstrom, Hedvig},
  booktitle=CVPR,
  pages={6158--6166},
  year={2017}
}

@inproceedings{cui2020learning,
  title={Learning dynamic relationships for 3d human motion prediction},
  author={Cui, Qiongjie and Sun, Huaijiang and Yang, Fei},
  booktitle=CVPR,
  pages={6519--6527},
  year={2020}
}

@inproceedings{jeong2024multi,
  title={Multi-agent long-term 3d human pose forecasting via interaction-aware trajectory conditioning},
  author={Jeong, Jaewoo and Park, Daehee and Yoon, Kuk-Jin},
  booktitle=CVPR,
  pages={1617--1628},
  year={2024}
}

@article{shu2021spatiotemporal,
  title={Spatiotemporal co-attention recurrent neural networks for human-skeleton motion prediction},
  author={Shu, Xiangbo and Zhang, Liyan and Qi, Guo-Jun and Liu, Wei and Tang, Jinhui},
  journal=PAMI,
  volume={44},
  number={6},
  pages={3300--3315},
  year={2021},
  publisher={IEEE}
}

@inproceedings{guo2022multi,
  title={Multi-person extreme motion prediction},
  author={Guo, Wen and Bie, Xiaoyu and Alameda-Pineda, Xavier and Moreno-Noguer, Francesc},
  booktitle=CVPR,
  pages={13053--13064},
  year={2022}
}

@inproceedings{chi3d,
  title={Three-dimensional reconstruction of human interactions},
  author={Fieraru, Mihai and Zanfir, Mihai and Oneata, Elisabeta and Popa, Alin-Ionut and Olaru, Vlad and Sminchisescu, Cristian},
  booktitle=CVPR,
  pages={7214--7223},
  year={2020}
}

@inproceedings{martinez2017human,
  title={On human motion prediction using recurrent neural networks},
  author={Martinez, Julieta and Black, Michael J and Romero, Javier},
  booktitle=CVPR,
  pages={2891--2900},
  year={2017}
}

@inproceedings{fang2023tbp,
  title={Tbp-former: Learning temporal bird's-eye-view pyramid for joint perception and prediction in vision-centric autonomous driving},
  author={Fang, Shaoheng and Wang, Zi and Zhong, Yiqi and Ge, Junhao and Chen, Siheng},
  booktitle=CVPR,
  pages={1368--1378},
  year={2023}
}

@inproceedings{gui2018adversarial,
  title={Adversarial geometry-aware human motion prediction},
  author={Gui, Liang-Yan and Wang, Yu-Xiong and Liang, Xiaodan and Moura, Jos{\'e} MF},
  booktitle=ECCV,
  pages={786--803},
  year={2018}
}

@article{zhuo2019unsupervised,
  title={Unsupervised online video object segmentation with motion property understanding},
  author={Zhuo, Tao and Cheng, Zhiyong and Zhang, Peng and Wong, Yongkang and Kankanhalli, Mohan},
  journal=TIP,
  volume={29},
  pages={237--249},
  year={2019},
  publisher={IEEE}
}

@article{shazeer2017outrageously,
  title={Outrageously large neural networks: The sparsely-gated mixture-of-experts layer},
  author={Shazeer, Noam and Mirhoseini, Azalia and Maziarz, Krzysztof and Davis, Andy and Le, Quoc and Hinton, Geoffrey and Dean, Jeff},
  journal={arXiv preprint arXiv:1701.06538},
  year={2017}
}

@article{pioro2024moe,
  title={Moe-mamba: Efficient selective state space models with mixture of experts},
  author={Pi{\'o}ro, Maciej and Ciebiera, Kamil and Kr{\'o}l, Krystian and Ludziejewski, Jan and Krutul, Micha{\l} and Krajewski, Jakub and Antoniak, Szymon and Mi{\l}o{\'s}, Piotr and Cygan, Marek and Jaszczur, Sebastian},
  journal={arXiv preprint arXiv:2401.04081},
  year={2024}
}

@article{lepikhin2020gshard,
  title={Gshard: Scaling giant models with conditional computation and automatic sharding},
  author={Lepikhin, Dmitry and Lee, HyoukJoong and Xu, Yuanzhong and Chen, Dehao and Firat, Orhan and Huang, Yanping and Krikun, Maxim and Shazeer, Noam and Chen, Zhifeng},
  journal={arXiv preprint arXiv:2006.16668},
  year={2020}
}

@article{fedus2022switch,
  title={Switch transformers: Scaling to trillion parameter models with simple and efficient sparsity},
  author={Fedus, William and Zoph, Barret and Shazeer, Noam},
  journal={JMLR},
  volume={23},
  number={120},
  pages={1--39},
  year={2022}
}

@inproceedings{vu2020anomaly,
  title={Anomaly detection in surveillance videos by future appearance-motion prediction},
  author={Vu, Tuan-Hung and Ambellouis, Sebastien and Boonaert, Jacques and Taleb-Ahmed, Abdelmalik},
  booktitle={ICCV Theory and Applications},
  pages={484--490},
  year={2020}
}

@article{tang2023collaborative,
  title={Collaborative uncertainty benefits multi-agent multi-modal trajectory forecasting},
  author={Tang, Bohan and Zhong, Yiqi and Xu, Chenxin and Wu, Wei-Tao and Neumann, Ulrich and Zhang, Ya and Chen, Siheng and Wang, Yanfeng},
  journal={IEEE Transactions on Pattern Analysis and Machine Intelligence},
  volume={45},
  number={11},
  pages={13297--13313},
  year={2023},
  publisher={IEEE}
}

@inproceedings{fragkiadaki2015recurrent,
  title={Recurrent network models for human dynamics},
  author={Fragkiadaki, Katerina and Levine, Sergey and Felsen, Panna and Malik, Jitendra},
  booktitle=ICCV,
  pages={4346--4354},
  year={2015}
}

@inproceedings{cui2021towards,
  title={Towards accurate 3d human motion prediction from incomplete observations},
  author={Cui, Qiongjie and Sun, Huaijiang},
  booktitle=CVPR,
  pages={4801--4810},
  year={2021}
}

@inproceedings{dang2021msr,
  title={Msr-gcn: Multi-scale residual graph convolution networks for human motion prediction},
  author={Dang, Lingwei and Nie, Yongwei and Long, Chengjiang and Zhang, Qing and Li, Guiqing},
  booktitle=CVPR,
  pages={11467--11476},
  year={2021}
}

@inproceedings{ma2022progressively,
  title={Progressively generating better initial guesses towards next stages for high-quality human motion prediction},
  author={Ma, Tiezheng and Nie, Yongwei and Long, Chengjiang and Zhang, Qing and Li, Guiqing},
  booktitle=CVPR,
  pages={6437--6446},
  year={2022}
}

@inproceedings{zhong2022spatio,
  title={Spatio-temporal gating-adjacency gcn for human motion prediction},
  author={Zhong, Chongyang and Hu, Lei and Zhang, Zihao and Ye, Yongjing and Xia, Shihong},
  booktitle=CVPR,
  pages={6447--6456},
  year={2022}
}

@inproceedings{mao2019learning,
  title={Learning trajectory dependencies for human motion prediction},
  author={Mao, Wei and Liu, Miaomiao and Salzmann, Mathieu and Li, Hongdong},
  booktitle=ICCV,
  pages={9489--9497},
  year={2019}
}

@inproceedings{mao2020history,
  title={History repeats itself: Human motion prediction via motion attention},
  author={Mao, Wei and Liu, Miaomiao and Salzmann, Mathieu},
  booktitle=ECCV,
  pages={474--489},
  year={2020}
}

@article{jacobs1991adaptive,
  title={Adaptive mixtures of local experts},
  author={Jacobs, Robert A and Jordan, Michael I and Nowlan, Steven J and Hinton, Geoffrey E},
  journal={Neural Comput.},
  volume={3},
  number={1},
  pages={79--87},
  year={1991},
  publisher={MIT Press}
}

@inproceedings{du2022glam,
  title={Glam: Efficient scaling of language models with mixture-of-experts},
  author={Du, Nan and Huang, Yanping and Dai, Andrew M and Tong, Simon and Lepikhin, Dmitry and Xu, Yuanzhong and Krikun, Maxim and Zhou, Yanqi and Yu, Adams Wei and Firat, Orhan and others},
  booktitle=ICML,
  pages={5547--5569},
  year={2022}
}

@article{liu2024deepseek,
  title={Deepseek-v3 technical report},
  author={Liu, Aixin and Feng, Bei and Xue, Bing and Wang, Bingxuan and Wu, Bochao and Lu, Chengda and Zhao, Chenggang and Deng, Chengqi and Zhang, Chenyu and Ruan, Chong and others},
  journal={arXiv preprint arXiv:2412.19437},
  year={2024}
}

@article{teamqwen1,
  title={Qwen1. 5-moe: Matching 7b model performance with 1/3 activated parameters},
  author={Team, Qwen},
  journal={https://qwenlm. github. io/blog/qwen-moe},
  year={2024}
}

@article{yang2025qwen3,
  title={Qwen3 technical report},
  author={Yang, An and Li, Anfeng and Yang, Baosong and Zhang, Beichen and Hui, Binyuan and Zheng, Bo and Yu, Bowen and Gao, Chang and Huang, Chengen and Lv, Chenxu and others},
  journal={arXiv preprint arXiv:2505.09388},
  year={2025}
}

@article{fu2022hungry,
  title={Hungry hungry hippos: Towards language modeling with state space models},
  author={Fu, Daniel Y and Dao, Tri and Saab, Khaled K and Thomas, Armin W and Rudra, Atri and R{\'e}, Christopher},
  journal={arXiv preprint arXiv:2212.14052},
  year={2022}
}

@article{gu2021efficiently,
  title={Efficiently modeling long sequences with structured state spaces},
  author={Gu, Albert and Goel, Karan and R{\'e}, Christopher},
  journal={arXiv preprint arXiv:2111.00396},
  year={2021}
}

@article{lin2025mgcn,
  title={MGCN: Mamba-integrated spatiotemporal graph convolutional network for long-term traffic forecasting},
  author={Lin, Wenxie and Zhang, Zhe and Ren, Gang and Zhao, Yangzhen and Ma, Jingfeng and Cao, Qi},
  journal={Knowl. Based Syst.},
  volume={309},
  pages={112875},
  year={2025}
}

@inproceedings{von2018recovering,
  title={Recovering accurate 3d human pose in the wild using imus and a moving camera},
  author={Von Marcard, Timo and Henschel, Roberto and Black, Michael J and Rosenhahn, Bodo and Pons-Moll, Gerard},
  booktitle=ECCV,
  pages={601--617},
  year={2018}
}

@inproceedings{mehta2018single,
  title={Single-shot multi-person 3d pose estimation from monocular rgb},
  author={Mehta, Dushyant and Sotnychenko, Oleksandr and Mueller, Franziska and Xu, Weipeng and Sridhar, Srinath and Pons-Moll, Gerard and Theobalt, Christian},
  booktitle={3DV},
  pages={120--130},
  year={2018}
}

@article{kingma2014adam,
  title={Adam: A method for stochastic optimization},
  author={Kingma, Diederik P and Ba, Jimmy},
  journal={arXiv preprint arXiv:1412.6980},
  year={2014}
}

@article{chu2021conditional,
  title={Conditional positional encodings for vision transformers},
  author={Chu, Xiangxiang and Tian, Zhi and Zhang, Bo and Wang, Xinlong and Shen, Chunhua},
  journal={arXiv preprint arXiv:2102.10882},
  year={2021}
}

@article{yun2024flex,
  title={Flex-moe: Modeling arbitrary modality combination via the flexible mixture-of-experts},
  author={Yun, Sukwon and Choi, Inyoung and Peng, Jie and Wu, Yangfan and Bao, Jingxuan and Zhang, Qiyiwen and Xin, Jiayi and Long, Qi and Chen, Tianlong},
  journal=NIPS,
  volume={37},
  pages={98782--98805},
  year={2024}
}

@article{li2024ftmomamba,
  title={FTMoMamba: Motion generation with frequency and text state space models},
  author={Li, Chengjian and Shu, Xiangbo and Cui, Qiongjie and Yao, Yazhou and Tang, Jinhui},
  journal={arXiv preprint arXiv:2411.17532},
  year={2024}
}

@article{qu2025mvp,
  title={MVP-shot: Multi-velocity progressive-alignment framework for few-shot action recognition},
  author={Qu, Hongyu and Yan, Rui and Shu, Xiangbo and Gao, Hailiang and Huang, Peng and Xie, Guo-Sen},
  journal=TMM,
  year={2025},
}

@article{qu2025learning,
  title={Learning clustering-based prototypes for compositional zero-shot learning},
  author={Qu, Hongyu and Wei, Jianan and Shu, Xiangbo and Wang, Wenguan},
  journal={arXiv preprint arXiv:2502.06501},
  year={2025}
}

@inproceedings{jiang2024delving,
  title={Delving into multimodal prompting for fine-grained visual classification},
  author={Jiang, Xin and Tang, Hao and Gao, Junyao and Du, Xiaoyu and He, Shengfeng and Li, Zechao},
  booktitle=AAAI,
  pages={2570--2578},
  year={2024}
}

@article{jiang2025fine,
  title={Fine-grained Image Retrieval via Dual-Vision Adaptation},
  author={Jiang, Xin and Cao, Meiqi and Tang, Hao and Shen, Fei and Li, Zechao},
  journal={arXiv preprint arXiv:2506.16273},
  year={2025}
}

@inproceedings{cao2025exploiting,
  title={Exploiting Frequency Dynamics for Enhanced Multimodal Event-based Action Recognition},
  author={Cao, Meiqi and Shu, Xiangbo and Jiang, Xin and Yan, Rui and Yao, Yazhou and Tang, Jinhui},
  booktitle=ICCV,
  pages={5969--5979},
  year={2025}
}

@inproceedings {xing2025vision,
  title={Vision-centric Token Compression in Large Language Model},
  author={Xing, Ling and Wang, Alex Jinpeng and Yan, Rui and Shu, Xiangbo and Tang, Jinhui},
  booktitle={NeurIPS},
  year={2025}
}

@inproceedings{qu2025omnigaze,
  title={OmniGaze: Reward-inspired Generalizable Gaze Estimation In The Wild},
  author={Qu, Hongyu and Wei, Jianan and Shu, Xiangbo and Yao, Yazhou and Wang, Wenguan and Tang, Jinhui},
  booktitle={NeurIPS},
  year={2025}
}

\appendix
\clearpage

\section{Appendix}
\setcounter{secnumdepth}{2} 

\section{Metric Definition}


\subsection{JPE Metric}
Across all experiments, we employ the mean per Joint Position Error (JPE) as the primary comparative metric. It is calculated as:
\begin{equation}
    \texttt{JPE}(\textbf{\textit{P}}_\text{pred}, \textbf{\textit{P}}_\text{gt}) = \frac{1}{J \cdot M} \sum_{i=1}^{M} \sum_{j=1}^{J} \| \hat{\textbf{\textit{P}}}_{j}^{i} - \textbf{\textit{P}}_{j}^{i} \|^2,
    \label{eq:jpe}
\end{equation}
where \(M\) and \(J\) are the numbers of people and joints. \(\hat{\textbf{\textit{P}}}_{j}^{i}\) and \(\textbf{\textit{P}}_{j}^{i}\) are the predicted and ground-truth positions of  the \(j\)-th joint for \(i\)-th person.  \(M\) and \(J\) are the numbers of people and joints. \(\hat{\textbf{\textit{P}}}_{j}^{i}\) and \(\textbf{\textit{P}}_{j}^{i}\) are the predicted and ground-truth positions of the \(j\)-th joint for \(i\)-th person. 

\subsection{APE Metric}
To isolate intrinsic pose errors, we remove global skeletal motion and quantify alignment accuracy using the Aligned Position Error (APE) metric. The calculation is as follows:
\begin{equation}
    \texttt{APE}(\textbf{\textit{P}}_\text{pred},\textbf{\textit{P}}_\text{gt}) = \texttt{JPE}(\textbf{\textit{P}}_\text{pred} - \textbf{\textit{P}}_\text{pred,r},\textbf{\textit{P}}_\text{gt} - \textbf{\textit{P}}_\text{gt,r}),
    \label{eq:ape}
\end{equation}
where \(\textbf{\textit{P}}_\text{pred,r}\) and \(\textbf{\textit{P}}_\text{gt,r}\) are the predicted and ground-truth positions of the root joint.





\section{Pseudocode}
\subsection{The Process of Spatiotemporal Mamba Block}
Algorithm~\ref{alg:smamba_block} and Algorithm~\ref{alg:tmamba_block} respectively illustrate the complete data processing procedures of spatial Mamba and temporal Mamba. Spatial Mamba blocks perform scanning along the pose dimension \(D\), while temporal Mamba blocks operate along the time dimension \(T\). Here \(B\) denotes the batch size and $N$ denotes SSM state expansion factor.
The Mamba Block initially expands the hidden dimension by a factor of $E$ via linear projection, yielding features $x$ and $z$. Subsequent convolutional operations and SiLU activation process $x$ to derive $x'$. An input-dependent selective SSM, which is central to the Mamba architecture, then discretizes system parameters ($\Delta, A, B$) and generates state representation $y$ from $x'$. This output is fused with the activated residual connection $\text{SiLU}(z)$ through element-wise multiplication, and then linearly transformed to yield the final time-step output $Y$.

\begin{algorithm}[t]
\caption{The process of spatial Mamba Block}
\label{alg:smamba_block}
\begin{algorithmic}[1]
\STATE \textbf{Input:} $\textbf{X} : (B, D, T)$  
\STATE \textbf{Output:} $\textbf{Y} : (B, D, T)$ 

\STATE $x, z : (B, D, ET) \leftarrow \text{Linear}(X)$ \COMMENT{Linear projection}
\STATE $x' : (B, D, ET) \leftarrow \text{SiLU}(\text{Conv1D}(x))$
\STATE $\textbf{A} : (T, N) \leftarrow \text{Parameter}$ \COMMENT{Structured state matrix}
\STATE $\textbf{B, C} : (B, D, N) \leftarrow \text{Linear}(x'), \text{Linear}(x')$
\STATE $\Delta : (B, D, T) \leftarrow \text{Softplus}(\text{Parameter} + \text{Broadcast}(\text{Linear}(x')))$
\STATE $\overline{\textbf{A}}, \overline{\textbf{B}} : (B, D, T, N) \leftarrow \text{discretize}(\Delta, \textbf{A}, \textbf{B})$ \COMMENT{Input-dependent parameters and discretization}
\STATE $y : (B, D, ET) \leftarrow \text{SelectiveSSM}(\overline{\textbf{A}}, \overline{\textbf{B}}, C)(x')$
\STATE $y' : (B, D, ET) \leftarrow y \otimes \text{SiLU}(z)$
\STATE $\textbf{Y} : (B, D, T) \leftarrow \text{Linear}(y')$ \COMMENT{Linear Projection}
\end{algorithmic}
\end{algorithm}

\begin{algorithm}[t]
\caption{The process of temporal Mamba Block}
\label{alg:tmamba_block}
\begin{algorithmic}[1]
\STATE \textbf{Input:} $\textbf{X} : (B, T, D)$  
\STATE \textbf{Output:} $\textbf{Y} : (B, T, D)$ 

\STATE $x, z : (B, T, ED) \leftarrow \text{Linear}(X)$ \COMMENT{Linear projection}
\STATE $x' : (B, T, ED) \leftarrow \text{SiLU}(\text{Conv1D}(x))$
\STATE $\textbf{A} : (D, N) \leftarrow \text{Parameter}$ \COMMENT{Structured state matrix}
\STATE $\textbf{B, C} : (B, T, N) \leftarrow \text{Linear}(x'), \text{Linear}(x')$
\STATE $\Delta : (B, T, D) \leftarrow \text{Softplus}(\text{Parameter} + \text{Broadcast}(\text{Linear}(x')))$
\STATE $\overline{\textbf{A}}, \overline{\textbf{B}} : (B, T, D, N) \leftarrow \text{discretize}(\Delta, \textbf{A}, \textbf{B})$ \COMMENT{Input-dependent parameters and discretization}
\STATE $y : (B, T, ED) \leftarrow \text{SelectiveSSM}(\overline{\textbf{A}}, \overline{\textbf{B}}, C)(x')$
\STATE $y' : (B, T, ED) \leftarrow y \otimes \text{SiLU}(z)$
\STATE $\textbf{Y} : (B, T, D) \leftarrow \text{Linear}(y')$ \COMMENT{Linear Projection}
\end{algorithmic}
\end{algorithm}

\section{Additional Ablation Studies}
\subsection{The Effectiveness of Bidirectional Scanning}

\begin{table}[t]
\centering
\setlength{\tabcolsep}{0.7mm} 
\small 
\begin{tabular}{c|cccc|cccc}
\hline
\multirow{2}{*}{Method} & \multicolumn{4}{c|}{JPE}                                       & \multicolumn{4}{c}{APE}                                        \\ \cline{2-9} 
                                   & 0.2s↓         & 0.6s↓         & 1.0s↓          & Avg↓          & 0.2s↓         & 0.6s↓         & 1.0s↓          & Avg↓          \\ \hline
forward                            & 33.7          & 99.8          & 164.4          & 99.3          & 23.7          & 73.7          & 105.2          & 67.5          \\ \hline
backward                           & 33.7          & 99.7          & 163.5          & 98.9          & 23.8          & 72.5          & 104.8          & 67.0          \\ \hline
\multicolumn{1}{l|}{bidirectional} & \textbf{31.4} & \textbf{95.3} & \textbf{158.3} & \textbf{95.0} & \textbf{22.1} & \textbf{70.4} & \textbf{103.8} & \textbf{65.4} \\ \hline
\end{tabular}%
\caption{Ablation study for effectiveness of bidirectional scanning strategy on CMU-Mocap (UMPM) dataset.}
\label{bidirectional}
\end{table}

To verify the global modeling capability of the bidirectional scanning strategy in the spatiotemporal experts, we replaced the bidirectional spatiotemporal Mamba with unidirectional spatiotemporal Mamba. As shown in Table \ref{bidirectional}, ``forward" indicates forward scanning, and ``backward" indicates backward scanning. We found that the results using either unidirectional scanning strategy are similar and lower than those obtained with bidirectional scanning. Specifically, using the bidirectional scanning strategy achieves an improvement of 4.3 mm and 3.9 mm in the average JPE metric, and an improvement of 2.1 mm and 1.6 mm in the average APE metric compared to the forward-only and backward-only scanning strategies, respectively. This is because the bidirectional scanning strategy possesses strong global sequence modeling capabilities, avoiding the limitations imposed by causal modeling.

\subsection{The Effectiveness of Mamba Architecture}

\begin{table}[t]
\centering
\setlength{\tabcolsep}{0.3mm} 
\small 
\begin{tabular}{c|cccc|cccc|l}
\hline
\multirow{2}{*}{Method} & \multicolumn{4}{c|}{JPE}                                       & \multicolumn{4}{c|}{APE}                                       & \multirow{2}{*}{{\begin{tabular}[c]{@{}c@{}}inference \\ speed\end{tabular}}} \\ \cline{2-9}
                        & 0.2s↓         & 0.6s↓         & 1.0s↓          & Avg↓          & 0.2s↓         & 0.6s↓         & 1.0s↓          & Avg↓          &                                 \\ \hline
MLP                     & 33.3          & 98.7          & 162.2          & 98.1          & 23.4          & 72.8          & 105.1          & 67.1          & 133ms/ite                       \\ \hline
Trans.             & 33.5          & 100.6         & 166.3          & 100.1         & 31.6          & 84.8          & 115.7          & 77.4         & 219ms/ite                       \\ \hline
Mamba                   & \textbf{31.4} & \textbf{95.3} & \textbf{158.3} & \textbf{95.0} & \textbf{22.1} & \textbf{70.4} & \textbf{103.8} & \textbf{65.4} & 171ms/ite                       \\ \hline
\end{tabular}%
\caption{Ablation study for effectiveness of Mamba architecture on CMU-Mocap (UMPM) dataset.}
\label{Mamba}
\end{table}

To verify the effectiveness of the Mamba architecture, we replaced the backbone of each spatiotemporal expert with MLP and Transformer, respectively. As shown in Table \ref{Mamba}, we found that the Mamba architecture achieves the best performance and offers better inference speed. Specifically, using the Mamba architecture achieves an improvement of 3.1 mm and 5.1 mm in the average JPE metric, and an improvement of 1.7 mm and 12 mm in the average APE metric compared to the MLP and Transformer architectures, respectively. Although models based on the MLP architecture can achieve faster inference speed, they fail to adequately capture spatiotemporal dependencies, resulting in suboptimal performance. In contrast, our Mamba architecture is particularly good at handling dependencies and thus achieves the best performance. Meanwhile, the Transformer-based architecture performs poorly due to the lack of spatiotemporal positional encoding, and it also achieves the slowest inference speed because of its quadratic time complexity.

\subsection{The Impact of Different Loss Weight Coefficient}

\begin{table}[t]
\centering
\begin{tabular}{c|c|c|c|c}
\hline
\(\alpha\) & \(\beta\)   & \(\lambda\)   & AVG JPE       & AVG APE       \\ \hline
1 & 1   & 1   & 97.4          & 66.9          \\ \hline
1 & 0   & 0.1 & 98.4          & 68.8          \\ \hline
1 & 0.1 & 0.1 & 97.7          & 68.1          \\ \hline
1 & 1   & 0   & 95.8          & 67.0          \\ \hline
1 & 1   & 0.1 & \textbf{95.0} & \textbf{65.4} \\ \hline
\end{tabular}%
\caption{The impact of different loss weight coefficient on CMU-Mocap (UMPM) dataset.}
\label{loss_coefficient}
\end{table}

To explore the impact of different weight coefficients, we conducted additional ablation experiments on the CMU-Mocap (UMPM) dataset. We set the coefficient for the predicted future joint spatial loss (\(\alpha\)) to 1, while setting the coefficients for temporal consistency loss (\(\beta\)) and historical joint spatial loss (\(\lambda\)) to various values. As shown in Table \ref{loss_coefficient}, we can observe that when the temporal consistency loss and historical joint position loss are removed, the average JPE metric decreases by 3.4 mm and 0.8 mm, respectively, indicating the importance of temporal consistency loss and historical joint position constraints. Conversely, when \(\alpha\) = \(\lambda\) = \(\beta\) = 1, the average JPE decreases by 2.4 mm, suggesting that overly constraining historical actions can lead the model to disregard its primary optimization goal, which is the future joint positions. The model achieves optimal performance when \(\alpha\) = \(\beta\) = 1 and \(\lambda\) = 0.1, highlighting that a well-balanced loss ratio is crucial for optimizing model performance.

\section{Additional Visualization Results}
\subsection{More Visualization on CMU-Mocap (UMPM)}
\begin{figure*}[t]
\centering
\includegraphics[width=1\textwidth]{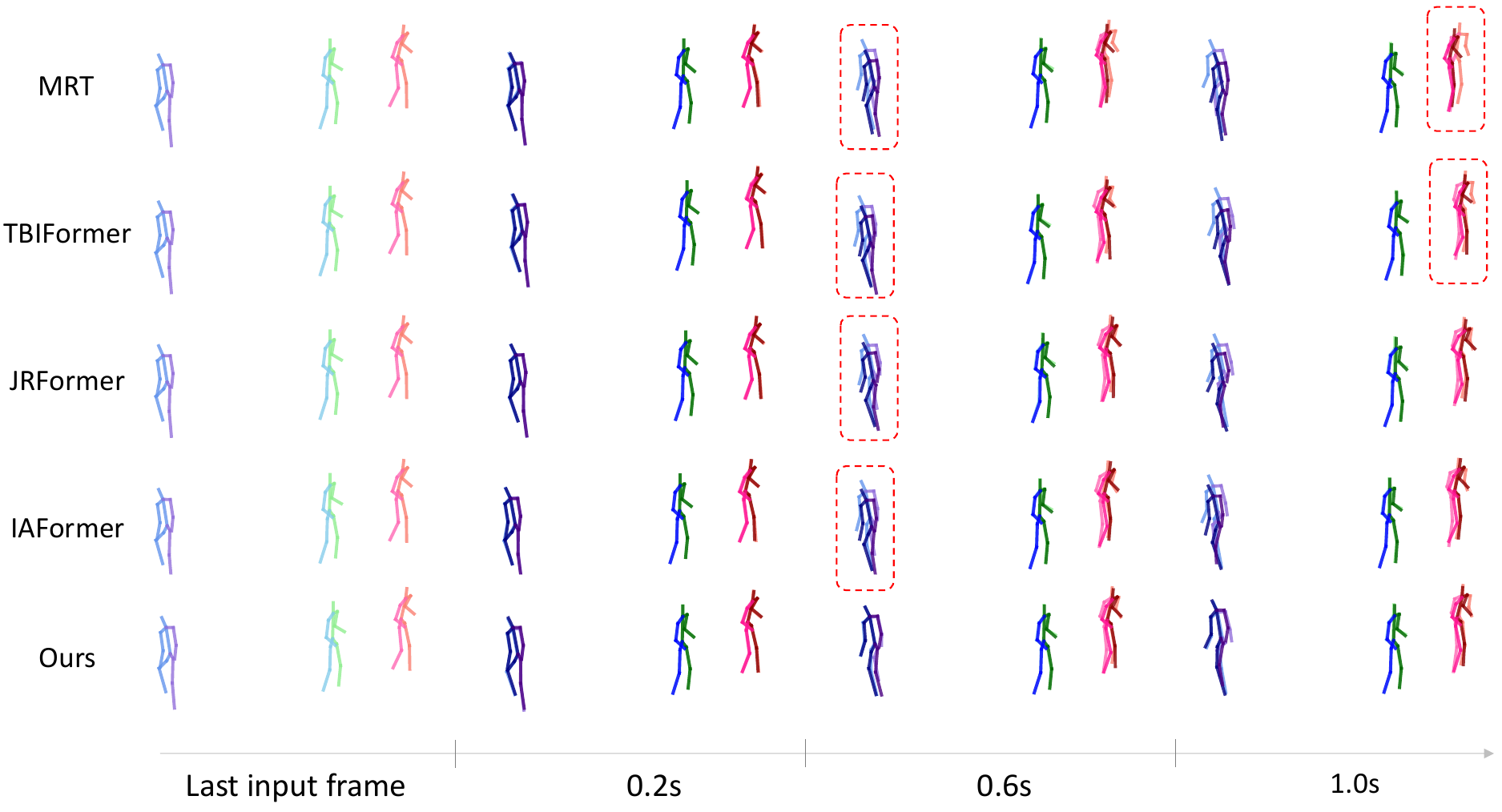} 
\caption{Additional visualization comparison on CMU-Mocap (UMPM) dataset. Dark-colored lines represent predicted motion, while light-colored lines indicate ground truth.}
\label{case1}
\end{figure*}

As shown in Fig. \ref{case1}, we compared the visual results of our method with those of MRT \cite{wang2021multi}, TBIFormer \cite{peng2023trajectory}, JRFormer \cite{xu2023joint}, and IAFormer \cite{xiao2024multi} on the CMU-Mocap (UMPM) dataset. The left column depicts the last input frames, while the three right columns show predictions at different timestamps. The individual on the left first turns right before beginning to ascend stairs, leading to a change in height. Meanwhile, the individual in the middle remains nearly stationary, and the one on the right moves backward. This example demonstrates the complex spatiotemporal dynamics exhibited by each individual. Our model adeptly captures these spatiotemporal dependencies through the utilization of four specialized spatiotemporal experts, yielding predictions that closely align with the ground truth. In contrast, existing methods struggle due to their reliance on fixed spatiotemporal position encoding, resulting in suboptimal performance.

\subsection{Qualitative Results with Video}
We have included videos of the visualization results in the multimedia appendix package to better observe the predictions of different models. The files are named as \texttt{video\_\{model\_name\}\_\{sample\_id\}.mp4}, where sample\_id 729 corresponds to the visualization results in the main paper, and sample\_id 1739 corresponds to additional visualization results in the supplementary material.

\section{Additional Remarks}
\subsection{More Results Analysis and Explanation}
We found that the prediction performance at 0.2s in the main paper is relatively poor, which may be because the model did not adequately capture the instantaneous dynamic changes within the prediction horizon. Specifically, we can subsequently try enabling the model to directly predict displacement vectors or state change quantities instead of absolute positions, to improve its sensitivity/perception to velocity changes. Furthermore, both TBIFormer and JRFormer integrated velocity information into their predictions, accounting for their stronger performance in short-term prediction. 

We found that adding the '+ST' component in Table \ref{table3} results in a significant performance degradation at the 0.2-second setting, and the potential reasons are as follows: "+ST" means all experts conduct spatial aggregation first, then temporal modeling. This "spatial-first" operation may distort short-term temporal cues and over-smooth joint high-frequency signals, while preserving refined joint positions and temporal continuity is critical for 0.2s short-term prediction.

\begin{table}[t]
\centering
\setlength{\tabcolsep}{0.7mm} 
\small 
\begin{tabular}{c|c|c|c}
\hline
Method        & \begin{tabular}[c]{@{}c@{}}Training \\ Speed↓ (ms/ite)\end{tabular} & \begin{tabular}[c]{@{}c@{}}Parameter \\ Size↓ (MB)\end{tabular} & \begin{tabular}[c]{@{}c@{}}AVG \\ JPE↓ (mm)\end{tabular} \\ \hline
MRT           & 174                                                                 & 7.27                                                            & 114                                                      \\ \hline
TBIFormer     & 1828                                                                & 6.70                                                            & 107                                                      \\ \hline
IAFormer      & 1134                                                                & 2.61                                                            & 96                                                       \\ \hline
\textbf{Ours} & \textbf{314}                                                        & \textbf{1.53}                                                   & \textbf{95}                                              \\ \hline
\end{tabular}%
\caption{Model efficiency comparison on CMU-Mocap (UMPM) dataset. For fair comparison, we train all models with the batch size of 96 using a single RTX 3090.}
\label{model_efficiency}
\end{table}

\subsection{More Dataset Details}
Following IAFormer \cite{xiao2024multi}, we employ the CHI3D dataset for model training and testing. Using folder s02 and s03 as the training set and s04 as the test set, we select 15 joints with indices 0, 1, 2, 3, 4, 5, 6, 8, 11, 12, 13, 14, 15, 16, and 9 to maintain consistency with mainstream multi-person motion prediction datasets. With a downsampling rate of 2, this yields a training set of 4,633 sequences and a test set of 1,208 sequences.

\subsection{Model Efficiency Comparison}
We present the comparative model efficiency of the main paper in tabular form, as shown in Table \ref{model_efficiency}. Our model achieves state-of-the-art performance while reducing the parameter count by 41.38\% and accelerating training by 3.6×.

\end{document}